%% file: main.tex
\documentclass[sigconf]{acmart}

\AtBeginDocument{%
  \providecommand\BibTeX{{%
    \normalfont B\kern-0.5em{\scshape i\kern-0.25em b}\kern-0.8em\TeX}}}

\copyrightyear{2022} 
\acmYear{2022} 
\setcopyright{acmcopyright}\acmConference[KDD '22]{Proceedings of the 28th ACM SIGKDD Conference on Knowledge Discovery and Data Mining}{August 14--18, 2022}{Washington, DC, USA}
\acmBooktitle{Proceedings of the 28th ACM SIGKDD Conference on Knowledge Discovery and Data Mining (KDD '22), August 14--18, 2022, Washington, DC, USA}
\acmPrice{15.00}
\acmDOI{10.1145/3534678.3539293}
\acmISBN{978-1-4503-9385-0/22/08}

\usepackage[ruled, lined, vlined, linesnumbered]{algorithm2e}
\usepackage{amsmath, amsthm}
\usepackage{arydshln}
\usepackage{bm}
\usepackage{comment}
\usepackage{multirow}
\usepackage{nccmath}
\usepackage{subfig}
\usepackage{threeparttable}
\usepackage{xcolor, xspace}

\theoremstyle{definition}

\newcommand{\eg}{\emph{e.g.},\xspace}
\newcommand{\ie}{\emph{i.e.},\xspace}

\newcommand\figref[1]{Fig.~\ref{#1}}
\newcommand\tabref[1]{Tab.~\ref{#1}}
\newcommand\secref[1]{Sec.~\ref{#1}}
\newcommand\equref[1]{Eq.(\ref{#1})}
\newcommand\algoref[1]{Alg.~\ref{#1}}

\newcommand{\fakeparagraph}[1]{\vspace{1mm}\noindent\textbf{#1.}}

\newcommand{\sysname}{p-Meta\xspace}

\begin{document}

\title{\sysname: Towards On-device Deep Model Adaptation}

\author{Zhongnan Qu}
\affiliation{%
  \institution{ETH Zurich}
  \city{Zurich}
  \country{Switzerland}}
\email{quz@ethz.ch}

\author{Zimu Zhou}
\affiliation{%
  \institution{Singapore Management University}
  \city{Singapore}
  \country{Singapore}}
\email{zimuzhou@smu.edu.sg}

\author{Yongxin Tong}
\affiliation{%
  \institution{Beihang University}
  \city{Beijing}
  \country{China}}
\email{yxtong@buaa.edu.cn}

\author{Lothar Thiele}
\affiliation{%
  \institution{ETH Zurich}
  \city{Zurich}
  \country{Switzerland}}
\email{thiele@ethz.ch}

\renewcommand{\shortauthors}{Zhongnan Qu et al.} 

\begin{abstract}
Data collected by IoT devices are often private and have a large diversity across users. Therefore, learning requires pre-training a model with available representative data samples, deploying the pre-trained model on IoT devices, and adapting the deployed model on the device with local data. Such an on-device adaption for deep learning empowered applications demands data and memory efficiency. However, existing gradient-based meta learning schemes fail to support memory-efficient adaptation. To this end, we propose p-Meta, a new meta learning method that enforces structure-wise partial parameter updates while ensuring fast generalization to unseen tasks. Evaluations on few-shot image classification and reinforcement learning tasks show that p-Meta not only improves the accuracy but also substantially reduces the peak dynamic memory by a factor of 2.5 on average compared to state-of-the-art few-shot adaptation methods.
\end{abstract}

%

\maketitle

\input{body/introduction}

\input{body/background}

\input{body/method}

\input{body/evaluation}

\input{body/related}

\input{body/conclusion}

\bibliographystyle{ACM-Reference-Format}
\bibliography{cites_short}

\clearpage
\input{body/appendix}

\end{document}

%% file: body/introduction.tex
\section{Introduction}
\label{sec:introduction}

Adaption to \textit{unseen} environments, users, and tasks is crucial for deep learning empowered IoT applications to deliver consistent performance and customized services. 
Data collected by IoT devices are often private and have a large diversity across users. 
For instance, activity recognition with smartphone sensors should adapt to countless walking patterns and sensor orientation \cite{bib:SenSys19:Gong}.
Human motion prediction with home robots needs fast learning of unseen poses for seamless human-robot interaction \cite{bib:ECCV18:Gui}.
In these applications, the \textit{new} data collected for model adaptation tend to relate to personal habits and lifestyle.
Hence, \textit{on-device} model adaptation is preferred over uploading the data to cloud servers for retraining.

Yet on-device adaption of a deep neural network (DNN) demands \textit{data efficiency} and \textit{memory efficiency}.
The excellent accuracy of contemporary DNNs is attributed to training with high-performance computers on large-scale datasets \cite{bib:book16:Goodfellow}.
For example, it takes $29$ hours to complete a $90$-epoch ResNet50 \cite{bib:CVPR16:He} training on ImageNet ($1.2$ million training images) \cite{bib:ILSVRC15} with $8$ NVIDIA Tesla P100 GPUs \cite{bib:arXiv17:Goyal}.
For on-device adaptation, however, neither abundant \textit{data} nor \textit{resources} are available.
A personal voice assistant, for example, may learn to adapt to users' accent and dialect within a few sentences, while a home robot should learn to recognize new object categories with few labelled images to navigate in new environments.
Furthermore, such adaptation is expected to be conducted on low-resource platforms such as smart portable devices, home hubs, and other IoT devices, with only several $KB$ to $MB$ memory.

\begin{figure}[t]
  \centering
  \includegraphics[width=0.42\textwidth]{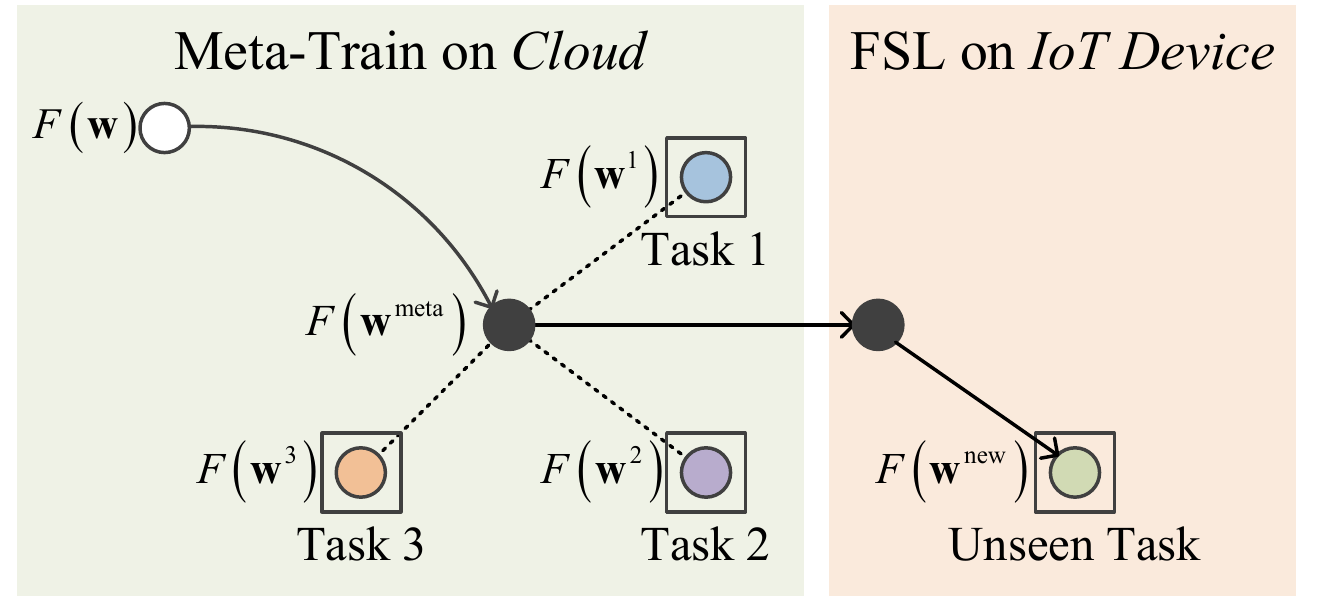} 
  \caption{Meta learning and few-shot learning (FSL) in the context of on-device adaptation. The backbone $F(\bm{w})$ is meta-trained into $F(\bm{w}^\mathrm{meta})$ on the cloud and is deployed to IoT devices to adapt to unseen tasks as $F(\bm{w}^\mathrm{new})$ via FSL.} \label{fig:meta}
  \vspace{-2em}
\end{figure}

For \textit{data-efficient} DNN adaptation, we resort to \textit{meta learning}, a paradigm that learns to fast generalize to unseen tasks \cite{bib:arXiv20:Hospedales}.
Of our particular interest is \textit{gradient-based} meta learning \cite{bib:ICLR19:Antreas, bib:ICML17:Finn, bib:ICLR20:Raghu, bib:ICLR21:Oh} for its wide applicability in classification, regression and reinforcement learning, as well as the availability of gradient-based training frameworks for low-resource devices, \eg TensorFlow Lite \cite{bib:tfliteOndeviceTraining}.
\figref{fig:meta} explains major terminologies in the context of on-device adaptation.
Given a backbone, its weights are \textit{meta-trained} on \textit{many} tasks, to output a \textit{model} that is expected to fast adapt to new \textit{unseen} tasks.
The process of adaptation is also known as \textit{few-shot learning}, where the meta-trained model is further retrained by standard stochastic gradient decent (SGD) on \textit{few new} samples only.

However, existing gradient-based meta learning schemes \cite{bib:ICLR19:Antreas, bib:ICML17:Finn, bib:ICLR20:Raghu, bib:ICLR21:Oh} fail to support \textit{memory-efficient} adaptation.
Although \textit{meta training} is conducted in the cloud, \textit{few-shot learning} (adaptation) of the meta-trained model is performed on IoT devices.
Consider to retrain a common backbone ResNet12 in a 5-way (5 new classes) 5-shot (5 samples per class) scenario.  
One round of SGD consumes 370.44MB peak dynamic memory, since the inputs of all layers must be stored to compute the gradients of these layers' weights in the backward path. 
In comparison, inference only needs 3.61MB.
The necessary dynamic memory is a key bottleneck for on-device adaptation due to cost and power constraints, even though the meta-trained model only needs to be retrained with a few data.

Prior efficient DNN training solutions mainly focus on parallel and distributed training on data centers \cite{bib:arXiv16:Chen, bib:ICLR21:Chen, bib:NIPS17:Greff, bib:NIPS16:Gruslys, bib:NIPS20:Raihan}.
On-device training has been explored for \textit{vanilla supervised training} \cite{bib:ICASSP20:Gooneratne, bib:MLSys21:Mathur, bib:SenSys19:Lee}, where training and testing are performed on the \textit{same} task. 
A pioneer study \cite{bib:NIPS20:Cai} investigated on-device adaptation to new tasks via memory-efficient \textit{transfer learning}. 
Yet transfer learning is prone to overfitting when only a few samples are available \cite{bib:ICML17:Finn}. 

In this paper, we propose \sysname, a new meta learning method for data- and memory-efficient DNN adaptation.
The key idea is to enforce \textit{structured partial parameter updates} while ensuring \textit{fast generalization to unseen tasks}. 
The idea is inspired by recent advances in understanding gradient-based meta learning \cite{bib:ICLR21:Oh, bib:ICLR20:Raghu}.
Empirical evidence shows that only the \textit{head} (the last output layer) of a DNN needs to be updated to achieve reasonable few-shot classification accuracy \cite{bib:ICLR20:Raghu} whereas the \textit{body} (the layers closed to the input) needs to be updated for cross-domain few-shot classification \cite{bib:ICLR21:Oh}.
These studies imply that certain weights are more important than others when generalizing to unseen tasks. Hence, we propose to automatically identify these \textit{adaptation-critical weights} to minimize the memory demand in few-shot learning.

Particularly, the critical weights are determined in two structured dimensionalities as, 
(\textit{i}) layer-wise: we meta-train a layer-by-layer learning rate that enables a \textit{static} selection of critical layers for updating; 
(\textit{ii}) channel-wise: we introduce meta attention modules in each layer to select critical channels \textit{dynamically}, \ie depending on samples from new tasks. 
Partial updating of weights means that (structurally) sparse gradients are generated, reducing memory requirements to those for computing nonzero gradients.
In addition, the computation demand for calculating zero gradients can be also saved.
To further reduce the memory, we utilize \textit{gradient accumulation} in few-shot learning and \textit{group normalization} in the backbone.
Although weight importance metrics and SGD with sparse gradients have been explored in vanilla training \cite{bib:NIPS20:Raihan, bib:PIEEE20:Deng, bib:ICASSP20:Gooneratne, bib:ICLR16:Han}, it is unknown \textit{(i)} how to identify adaptation-critical weights and \textit{(ii)} whether meta learning is robust to sparse gradients, where the objective is fast adaptation to \textit{unseen} tasks.

Our main contributions are summarized as follows.
\begin{itemize}
    \item 
    We design \sysname, a new meta learning method for data- and memory-efficient DNN adaptation to unseen tasks.
    \sysname automatically identifies adaptation-critical weights both layer-wise and channel-wise for low-memory adaptation. 
    The hierarchical approach combines static identification of layers and dynamic identification of channels whose weights are critical for few-shot adaptation.
    To the best of our knowledge, \sysname is the first meta learning method designed for on-device few-shot learning.
    \item
    Evaluations on few-shot image classification and reinforcement learning show that, \sysname not only improves the accuracy but also reduces the peak dynamic memory by a factor of 2.5 on average over the state-of-the-art few-shot adaptation methods. 
    \sysname can also simultaneously reduce the computation by a factor of 1.7 on average.
\end{itemize}

In the rest of this paper, we introduce the preliminaries and challenges in \secref{sec:background}, elaborate on the design of \sysname in \secref{sec:method}, present its evaluations in \secref{sec:evaluation}, review related work in \secref{sec:related}, and conclude in \secref{sec:conclusion}.

%% file: body/background.tex
\section{Preliminaries and Challenges}
\label{sec:background}
In this section, we provide the basics on meta learning for fast adaptation and highlight the challenges to enable on-device adaptation.

\fakeparagraph{Meta Learning for Fast Adaptation}
Meta learning is a prevailing solution to adapt a DNN to unseen tasks with limited training samples, \ie few-shot learning \cite{bib:arXiv20:Hospedales}.
We ground our work on model-agnostic meta learning (MAML) \cite{bib:ICML17:Finn}, a generic meta learning framework which supports classification, regression and reinforcement learning.
Given the dataset $\mathcal{D}=\{\mathcal{S}, \mathcal{Q}\}$ of an unseen few-shot task, where $\mathcal{S}$ (support set) and $\mathcal{Q}$ (query set) are for training and testing, MAML trains a model $F(\bm{w})$ with weights $\bm{w}$ such that it yields high accuracy on $\mathcal{Q}$ even when $\mathcal{S}$ only contains a few samples.
This is enabled by simulating the few-shot learning experiences over abundant few-shot tasks sampled from a task distribution $p(\mathsf{T})$.
Specifically, it meta-trains a backbone $F$ over few-shot tasks $\mathsf{T}^{i}\sim p(\mathsf{T})$, where each $\mathsf{T}^{i}$ has dataset $\mathcal{D}^i=\{\mathcal{S}^i,\mathcal{Q}^i\}$, and then generates $F(\bm{w}^\mathrm{meta})$, an initialization for the unseen few-shot task $\mathsf{T}^{\mathrm{new}}$ with dataset $\mathcal{D}^{\mathrm{new}}=\{\mathcal{S}^{\mathrm{new}},\mathcal{Q}^{\mathrm{new}}\}$.
Training from $F(\bm{w}^\mathrm{meta})$ over $\mathcal{S}^{\mathrm{new}}$ is expected to achieve a high test accuracy on $\mathcal{Q}^{\mathrm{new}}$.

MAML achieves fast adaptation via two-tier optimization. 
In the \textit{inner loop}, a task $\mathsf{T}^i$ and its dataset $\mathcal{D}^i$ are sampled. 
The weights $\bm{w}$ are updated to $\bm{w}^{i}$ on support dataset $\mathcal{S}^{i}$ via $K$ gradient descent steps, where $K$ is usually small, compared to vanilla training:
\begin{equation}\label{eq:maml_inner}
    \medop{\bm{w}^{i,k} = \bm{w}^{i,k-1}-\alpha\nabla_{\bm{w}}~\ell\left(\bm{w}^{i,k-1};\mathcal{S}^i\right)\quad\mathrm{for}~k=1\cdots K}
\end{equation}
where $\bm{w}^{i,k}$ are the weights at step $k$ in the inner loop, and $\alpha$ is the inner step size.
Note that $\bm{w}^{i,0}=\bm{w}$ and $\bm{w}^{i}=\bm{w}^{i,K}$.
$\ell(\bm{w};\mathcal{D})$ is the loss function on dataset $\mathcal{D}$. 
In the \textit{outer loop}, the weights are optimized to minimize the sum of loss at $\bm{w}^{i}$ on query dataset $\mathcal{Q}^i$ across tasks. 
The gradients to update weights in the outer loop are calculated w.r.t. the starting point $\bm{w}$ of the inner loop.
\begin{equation}\label{eq:maml_outer}
    \medop{\bm{w} \leftarrow \bm{w}-\beta \nabla_{\bm{w}} \sum_i \ell\left(\bm{w}^{i};\mathcal{Q}^i\right)}
\end{equation}
where $\beta$ is the outer step size.

\begin{table*}[t]
    \centering
    \caption{The static memory, the peak dynamic memory, and the total computation (GMACs $=10^9$MACs) of inference and adaptation for sample applications. For image classification we use batch size $=25$. For robot locomotion we use rollouts $=20$, horizon $=200$; each sample corresponds to a rollouted episode, and the case for an observation is reported in brackets. The calculation is based on Appendix \ref{app:analysis}.}
    \label{tab:examples}
    \vspace{-0.2em}
    \footnotesize
    \begin{tabular}{llcccccc}
    	\toprule
    	\multirow{2}{*}{Application}    & \multirow{2}{*}{Model / Benchmark} & \multicolumn{2}{c}{Static Memory (MB)} & \multicolumn{2}{c}{Peak Dynamic Memory (MB)} & \multicolumn{2}{c}{GMACs}\\
    	\cmidrule(lr){3-4} \cmidrule(lr){5-6} \cmidrule(lr){7-8} & & Model & Sample & Inference & Adaptation & Inference & Adaptation \\
    	\midrule
    	Image Classification & 4Conv \cite{bib:ICML17:Finn} / MiniImageNet \cite{bib:NIPS16:Vinyals} & $0.13$ & $0.53$ & $0.90$ & $48.33$ & $0.72$ & $1.96$ \\
    	Image Classification & ResNet12 \cite{bib:NIPS18:Oreshkin} / MiniImageNet \cite{bib:NIPS16:Vinyals} & $32.0$ & $0.53$ & $3.61$ & $370.44$ & $62.08$ & $185.42$ \\
    	Robot Locomotion & MLP \cite{bib:ICML17:Finn} / MuJoCo \cite{bib:IROS12:Todorov} & $0.05$ & $0.016 (0.00008)$ & $0.08 (0.0004)$ & $3.72$  & $0.05$ & $0.15$ \\
    	\bottomrule
    \end{tabular}
\end{table*}
 
The meta-trained weights $\bm{w}^\mathrm{meta}$ are then used as initialization for few-shot learning into $\bm{w}^\mathrm{new}$ by $K$ gradient descent steps over $\mathcal{S}^{\mathrm{new}}$.
Finally we assess the accuracy of $F(\bm{w}^\mathrm{new})$ on $\mathcal{Q}^{\mathrm{new}}$.

\fakeparagraph{Memory Bottleneck of On-device Adaptation}
As mentioned above, the meta-trained model $F(\bm{w}^\mathrm{meta})$ can adapt to unseen tasks via $K$ gradient descent steps.
Each step is the same as the inner loop of meta-training \equref{eq:maml_inner}, but on dataset $\mathcal{S}^{\mathrm{new}}$.
\begin{equation}\label{eq:fsl}
    \medop{\bm{w}^{\mathrm{new},k} = \bm{w}^{\mathrm{new},k-1}-\alpha\nabla_{\bm{w}^{\mathrm{new}}}~\ell\left(\bm{w}^{\mathrm{new},k-1};\mathcal{S}^{\mathrm{new}}\right)}
\end{equation}
where $\bm{w}^{\mathrm{new},0}=\bm{w}^\mathrm{meta}$.
For brevity, we omit the superscripts of model adaption in \equref{eq:fsl} and use $\bm{g}(\cdot)$ as the loss gradients w.r.t. the given tensor.
Hence, without ambiguity, we simplify the notations of \equref{eq:fsl} as follows:
\begin{equation}\label{eq:fsl_s}
    \medop{\bm{w} \leftarrow \bm{w}-\alpha \bm{g}(\bm{w})}
\end{equation}

\begin{figure}[t]
  \centering
  \includegraphics[width=0.4\textwidth]{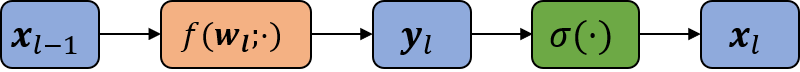} 
  \caption{A typical layer $l$ in DNNs. $\bm{x}_{l-1}$ is the input tensor; $\bm{x}_l$ is the output tensor, also the input tensor of layer $l+1$; $\bm{y}_l$ is the intermediate tensor; $\bm{w}_l$ is the weight tensor. }
\label{fig:layer}
\end{figure}

Let us now understand where the main memory cost for iterating \equref{eq:fsl_s} comes from. 
For the sake of clarity, we focus on a feed forward DNNs that consist of $L$ convolutional (\texttt{conv}) layers or fully-connected (\texttt{fc}) layers.
A typical layer (see \figref{fig:layer}) consists of two operations: (\textit{i}) a linear operation with trainable parameters, \eg convolution or affine; (\textit{ii}) a parameter-free non-linear operation (may not exist in certain layers), where we consider max-pooling or ReLU-styled (ReLU, LeakyReLU) activation functions in this paper.

Take a network consisting of \texttt{conv} layers only as an example. 
The memory requirements for storing the activations $\bm{x}_l\in\mathbb{R}^{C_l\times H_l \times W_l}$ as well as the convolution weights $\bm{w}_l \in \mathbb{R}^{C_l\times C_{l-1}\times S_l \times S_l} $ of layer $l$ in words can be determined as
\[
\medop{m(\bm{x}_l)= C_l H_l W_l \; , \; \; m(\bm{w}_l) = C_l C_{l-1} S_l^2}
\]
where $C_{l-1}$, $C_l$, $H_l$, and $W_l$ stand for input channel number, output channel number, height and width of layer $l$, respectively; $S_l$ stands for the kernel size.
The detailed memory and computation demand analysis as provided in Appendix \ref{app:analysis} reveals that the by far largest memory requirement is neither attributed to determining the activations $\bm{x}_l$ in the forward path nor to determining the gradients of the activations $\bm{g}(\bm{x}_l)$ in the backward path. 
Instead, the memory bottleneck lies in the computation of the weight gradients $\bm{g}(\bm{w}_l)$, which requires the availability of the activations $\bm{x}_{l-1}$ from the forward path. 
Following \equref{eq:memorySimple} in Appendix \ref{app:analysis}, the necessary memory in words is
\begin{equation}
\label{eq:MemoryContrib}
\medop{\sum_{1 \leq l \leq L} m(\bm{x}_{l-1})}
\end{equation}

\tabref{tab:examples} summarizes the memory consumption and the total computation of the commonly used few-shot learning backbone models \cite{bib:ICML17:Finn, bib:NIPS18:Oreshkin}. 
The requirements are based on the detailed analysis in Appendix~\ref{app:analysis}.  
We can draw two intermediate conclusions. 
\begin{itemize}
    \item 
    The total computation of adaptation (training) is approximately $2.7\times$ to $3\times$ larger compared to inference.
    Yet the peak dynamic memory of training is far larger, $47 \times$ to $103 \times$ over inference. 
    The peak dynamic memory consumption of training is also also significantly higher than the static memory consumption from the model and the training samples in few-shot learning. 
    \item
    To enable adaptation for memory-constrained IoT devices, we need to find some way of getting rid of the major dynamic memory contribution in \equref{eq:MemoryContrib}.
\end{itemize}

%% file: body/method.tex
\section{Method}\label{sec:method}
This section presents \sysname, a new meta learning scheme that enables memory-efficient few-shot learning on unseen tasks.

\subsection{\sysname Overview}
\label{sec:overview}
We first provide an overview of \sysname and introduce its main concepts, namely selecting critical gradients, using a hierarchical approach to determine adaption-critical layers and channels, and using a mixture of static and dynamic selection mechanisms. 

\fakeparagraph{Principles}
We impose \textit{structured sparsity} on the \textit{gradients} $\bm{g}(\bm{w}_l)$ such that the corresponding tensor dimensions of $\bm{x}_l$ do not need to be saved. 
There are other options to reduce the dominant memory demand in \equref{eq:MemoryContrib}.
They are inapplicable for the reasons below.
\begin{itemize}
    \item 
    One may trade-off computation and memory by recomputing activations $\bm{x}_{l-1}$ when needed for determining $\bm{w}_l$, see for example \cite{bib:arXiv16:Chen, bib:NIPS16:Gruslys}. 
    Due to the limited processing abilities of IoT devices, we exclude this option. 
    \item
    It is also possible to prune activations $\bm{x}_{l-1}$. 
    Yet based on our experiments in Appendix \ref{app:sparse}, imposing sparsity on $\bm{x}_{l-1}$ hugely degrades few-shot learning accuracy as this causes error accumulation along the propagation, see also \cite{bib:NIPS20:Raihan}.
    \item
    Note that unstructured sparsity, as proposed in \cite{bib:CIKM21:Gao, bib:NIPS21:Oswald}, does not in general lead to memory savings, since there is a very small probability that all weight gradients for which an element of $\bm{x}_{l-1}$ is necessary have been pruned. 
\end{itemize}

\noindent
We impose sparsity on the gradients in a hierarchical manner.

\begin{itemize}
    \item 
    \textbf{Selecting adaption-critical layers.} 
    We first impose layer-by-layer sparsity on $\bm{g}(\bm{w}_l)$.
    It is motivated by previous results showing that manual freezing of certain layers does no harm to few-shot learning accuracy \cite{bib:ICLR20:Raghu,bib:ICLR21:Oh}. 
    Layer-wise sparsity reduces the number of layers whose weights need to be updated. 
    We determine the adaptation-critical layers from the meta-trained \textit{layer-wise sparse learning rates}. 
    
    \item
    \textbf{Selecting adaption-critical channels.}
    We further reduce the memory demand by imposing sparsity on $\bm{g}(\bm{w}_l)$ within each layer. Noting that calculating $\bm{g}(\bm{w}_l)$ needs both the input channels $\bm{x}_{l-1}$ and the output channels $\bm{g}(\bm{y}_{l})$, we enforce sparsity on both of them.
    Input channel sparsity decreases memory and computation overhead, whereas output channel sparsity improves few-shot learning accuracy and reduces computation.
    We design a novel \textit{meta attention mechanism} to \textit{dynamically} determine adaptation-critical channels. 
    They take as inputs $\bm{x}_{l-1}$ and $\bm{g}(\bm{y}_{l})$ and determine adaptation-critical channels during few-shot learning, based on the given few data samples from new unseen tasks. 
    Dynamic channel-wise learning rates as determined by meta attention yield a significantly higher accuracy than a static channel-wise learning rate (see \secref{sec:ablation}).
\end{itemize}

\fakeparagraph{Memory Reduction}  
The reduced memory demand due to our hierarchical approach can be seen in \equref{eq:memorySimple} in Appendix \ref{app:analysis}:
\begin{equation*}
\medop{\sum_{1 \leq l \leq L} \hat{\alpha}_l \mu^{\mathrm{fw}}_l m(\bm{x}_{l-1})}
\end{equation*}
where $\hat{\alpha}_l \in \{ 0, 1\}$ is the mask from the static selection of critical layers and $0 \leq \mu^{\mathrm{fw}}_l \leq 1$ denotes the relative amount of dynamically chosen input channels.

Next, we explain how \sysname selects adaptation-critical layers (\secref{sec:LR}) and channels within layers (\secref{sec:attention}) as well as the deployment optimizations (\secref{sec:others}) for memory-efficient adaptation.

\subsection{Selecting Adaption-Critical Layers by Learning Sparse Inner Step Sizes}
\label{sec:LR}
This subsection introduces how \sysname meta-learns adaptation-critical layers to reduce the number of updated layers during few-shot learning.
Particularly, instead of manual configuration as in \cite{bib:ICLR21:Oh, bib:ICLR20:Raghu}, we propose to automate the layer selection process.
During meta training, we identify adaptation-critical layers by learning layer-wise sparse inner step sizes (\secref{subsec:LR:ml}).
Only these critical layers with nonzero step sizes will be updated during on-device adaptation to new tasks (\secref{subsec:LR:fsl}).

\subsubsection{Learning Sparse Inner Step Sizes in Meta Training}
\label{subsec:LR:ml}

Prior work \cite{bib:ICLR19:Antreas} suggests that instead of a global fixed inner step size $\alpha$, learning the inner step sizes $\bm{\alpha}$ for each layer and each gradient descent step improves the generalization of meta learning, where $\bm{\alpha} = \alpha_{1:L}^{1:K} \succeq \bm{0}$.
We utilize such learned inner step sizes to infer layer importance for adaptation.
We learn the inner step sizes $\bm{\alpha}$ in the outer loop of meta-training while fixing them in the inner loop.

\fakeparagraph{Learning Layer-wise Inner Step Sizes}
We change the inner loop of \equref{eq:maml_inner} to incorporate the per-layer inner step sizes:
\begin{equation}\label{eq:inner_alpha}
    \medop{\bm{w}_l^{i,k} = \bm{w}_l^{i,k-1}-\alpha^k_l\nabla_{\bm{w}_l}~\ell\left(\bm{w}^{i,k-1}_{1:L};\mathcal{S}^i\right)}
\end{equation}
where $\bm{w}_l^{i,k}$ is the weights of layer $l$ at step $k$ optimized on task $i$ (dataset $\mathcal{S}^{i}$).
In the outer loop, weights $\bm{w}$ are still optimized as
\begin{equation}\label{eq:outer_alpha}
    \medop{\bm{w} \leftarrow \bm{w}-\beta \nabla_{\bm{w}} \sum_i \ell\left(\bm{w}^{i};\mathcal{Q}^i\right)}
\end{equation}
where $\bm{w}^{i}=\bm{w}^{i,K}=\bm{w}^{i,K}_{1:L}$, which is a function of $\bm{\alpha}$.
The inner step sizes $\bm{\alpha}$ are then optimized as
\begin{equation}\label{eq:alpha}
    \medop{\bm{\alpha} \leftarrow \bm{\alpha}-\beta \nabla_{\bm{\alpha}} \sum_i \ell\left(\bm{w}^{i};\mathcal{Q}^i\right)}
\end{equation}

\fakeparagraph{Imposing Sparsity on Inner Step Sizes}
To facilitate layer selection, we enforce sparsity in $\bm{\alpha}$, \ie encouraging a subset of layers to be selected for updating.
Specifically, we add a Lasso regularization term in the loss function of \equref{eq:alpha} when optimizing $\bm{\alpha}$.
Hence, the final optimization of $\bm{\alpha}$ in the outer loop is formulated as
\begin{equation}\label{eq:regularization}
    \medop{\bm{\alpha} \leftarrow \bm{\alpha}-\beta \nabla_{\bm{\alpha}} (\sum_i \ell\left(\bm{w}^{i};\mathcal{Q}^i\right)+\lambda \sum_{l,k} m(\bm{x}_{l-1})\cdot|\alpha_l^k|)}
\end{equation}
where $\lambda$ is a positive scalar to control the ratio between two terms in the loss function.
We empirically set $\lambda=0.001$.
$|\alpha_l^k|$ is re-weighted by $m(\bm{x}_{l-1})$, which denotes the necessary memory in \equref{eq:MemoryContrib} if only updating the weights in layer $l$. 

\subsubsection{Exploiting Sparse Inner Step Sizes for On-device Adaptation}
\label{subsec:LR:fsl}
We now explain how to apply the learned $\bm{\alpha}$ to save memory during on-device adaptation.
After deploying the meta-trained model to IoT devices for adaptation, at updating step $k$, for layers with $\alpha_l^k=0$, the activations (\ie their inputs) $\bm{x}_{l-1}$ need not be stored, see \equref{eq:memoryAll} and \equref{eq:memorySimple} in Appendix~\ref{app:analysis}. 
In addition, we do not need to calculate the corresponding weight gradients $\bm{g}(\bm{w}_l)$, which saves computation, see \equref{eq:MAC} in Appendix~\ref{app:analysis}.

\subsection{Selecting Adaption-Critical Channels within Layers via Sparse Meta Attention}
\label{sec:attention}

\begin{figure}[t]
\vspace{-0.5em}
  \centering
  \includegraphics[width=0.5\textwidth]{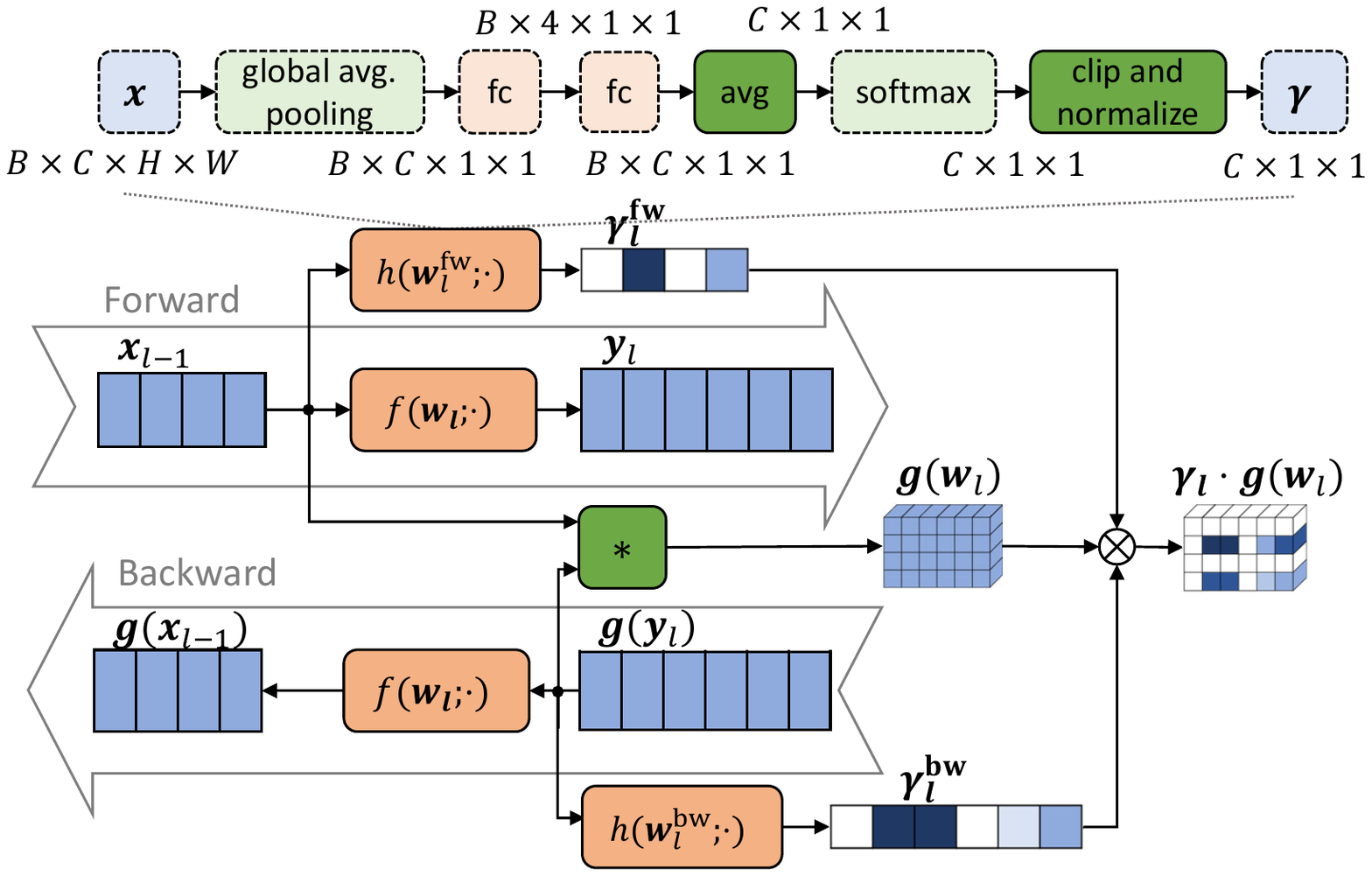} 
  \vspace{-2.5em}
  \caption{Meta attention of layer $l$ during meta-training. The blue blocks correspond to tensors; the orange blocks correspond to computation units with parameters, and the green ones without. Each column of a tensor corresponds to one channel. The input tensor $\bm{x}_{l-1}$ has 4 channels; the output tensor $\bm{y}_{l}$ has 6 channels. The other dimensions (\eg height, width and batch) are omitted here. The green block with $*$ stands for the operations involved to compute $\bm{g}(\bm{w}_l)$. In order to compute the gradients of the parameters in meta attention, \ie $\bm{w}_l^{\mathrm{fw}}$ and $\bm{w}_l^{\mathrm{bw}}$, the full dense gradients $\bm{g}(\bm{w}_l)$ are computed during meta-training, and then are masked by $\bm{\gamma}_l$. An example meta attention module for a \texttt{conv} layer is shown in the upper part. $B$ denotes the batch size. The newly added blocks related to the inference attention in \cite{bib:CVPR20:Chen} are marked with solid lines.}
\label{fig:metaattention}
\end{figure}

This subsection explains how \sysname learns a novel meta attention mechanism in each layer to dynamically select adaptation-critical channels for further memory saving in few-shot learning.
Despite the widespread adoption of channel-wise attention for inference \cite{bib:CVPR18:Hu, bib:CVPR20:Chen}, we make the first attempt to use attention for memory-efficient training (few-shot learning in our case).
For each layer, its meta attention outputs a dynamic channel-wise sparse attention score based on the samples from new tasks.
The sparse attention score is used to re-weight (also sparsify) the weight gradients.
Therefore, by calculating only the nonzero gradients of critical weights within a layer, we can save both memory and computation.
We first present our meta attention mechanism during meta training (\secref{sec:attention:ml}) and then show its usage for on-device model adaptation (\secref{sec:attention:fsl}).

\subsubsection{Learning Sparse Meta Attention in Meta Training}
\label{sec:attention:ml}
Since mainstream backbones in meta learning use small kernel sizes (1 or 3), we design the meta attention mechanism channel-wise.
\figref{fig:metaattention} illustrates the attention design during meta-training.

\fakeparagraph{Learning Meta Attention}
The attention mechanism is as follows.
\begin{itemize}
    \item 
    We assign an attention score to the weight gradients of layer $l$ in the inner loop of meta training.
    The attention scores are expected to indicate which weights/channels are important and thus should be updated in layer $l$.
    \item
    The attention score is obtained from two attention modules: one taking $\bm{x}_{l-1}$ as input in the forward pass, and the other taking $\bm{g}(\bm{y}_l)$ as input during the backward pass.
    We use $\bm{x}_{l-1}$ and $\bm{g}(\bm{y}_l)$ to calculate the attention scores because they are used to compute the weight gradients $\bm{g}(\bm{w}_{l})$.
\end{itemize}
Concretely, we define the forward and backward attention scores for a \texttt{conv} layer as,
\begin{equation}\label{eq:atten_fw}
    \medop{\bm{\gamma}^{\mathrm{fw}}_{l} = h(\bm{w}^{\mathrm{fw}}_l;\bm{x}_{l-1})\in\mathbb{R}^{C_{l-1}\times 1 \times 1}}
\end{equation}
\begin{equation}\label{eq:atten_bw}
    \medop{\bm{\gamma}^{\mathrm{bw}}_{l} = h(\bm{w}^{\mathrm{bw}}_l;\bm{g}(\bm{y}_l))\in\mathbb{R}^{C_{l}\times 1 \times 1}}
\end{equation}
where $h(\cdot;\cdot)$ stands for the meta attention module, and $\bm{w}^{\mathrm{fw}}_l$ and $\bm{w}^{\mathrm{bw}}_l$ are the parameters of the meta attention modules.
The overall (sparse) attention scores $\bm{\gamma}_l\in\mathbb{R}^{C_{l}\times C_{l-1} \times 1 \times 1}$ and is computed as, 
\begin{equation}\label{eq:atten_fwbw}
    \medop{\gamma_{l,ba11} = \gamma^{\mathrm{fw}}_{l,a11} \cdot \gamma^{\mathrm{bw}}_{l,b11}} 
\end{equation}
In the inner loop, for layer $l$, step $k$ and task $i$, $\bm{\gamma}_l$ is (broadcasting) multiplied with the dense weight gradients to get the sparse ones,
\begin{equation}\label{eq:inner_incr}
    \medop{\bm{\gamma}_l^{i,k}\odot\nabla_{\bm{w}_l}~\ell\left(\bm{w}^{i,k-1}_{1:L};\mathcal{S}^i\right)}
\end{equation}
The weights are then updated by,
\begin{equation}\label{eq:inner_atten}
    \medop{\bm{w}_l^{i,k} = \bm{w}_l^{i,k-1}-\alpha^k_l(\bm{\gamma}_l^{i,k}\odot\nabla_{\bm{w}_l}~\ell\left(\bm{w}^{i,k-1}_{1:L};\mathcal{S}^i\right))}
\end{equation}
Let all attention parameters be $\bm{w}^{\mathrm{atten}} = \{\bm{w}^{\mathrm{fw}}_l,\bm{w}^{\mathrm{bw}}_l\}_{l=1}^{L}$. 
The attention parameters $\bm{w}^{\mathrm{atten}}$ are optimized in the outer loop as,
\begin{equation}\label{eq:outer_atten}
    \medop{\bm{w}^{\mathrm{atten}} \leftarrow \bm{w}^{\mathrm{atten}}-\beta \nabla_{\bm{w}^{\mathrm{atten}}} \sum_i \ell\left(\bm{w}^{i};\mathcal{Q}^i\right)}
\end{equation}
Note that we use a dense forward path and a dense backward path in both meta-training and on-device adaptation, as shown in \figref{fig:metaattention}.
That is, the attention scores $\bm{\gamma}^{\mathrm{fw}}_{l}$ and $\bm{\gamma}^{\mathrm{bw}}_{l}$ are only calculated locally and will not affect $\bm{y}_l$ during forward and $\bm{g}(\bm{x}_{l-1})$ during backward.

\begin{algorithm}[t]
\caption{Clip and normalization}\label{alg:clip}
\KwIn{softmax output (normalized) $\bm{\pi}\in\mathbb{R}^{C}$, clip ratio $\rho$} 
\KwOut{sparse $\bm{\gamma}$}
Sort $\bm{\pi}$ in ascending order and get sorted indices $d_{1:C}$\; 
Find the smallest $c$ such that $\sum_{i=1}^c\pi_{d_i}\ge\rho$\;
Set $\pi_{d_1:d_c}$ as 0 \tcp*{if $\rho=0$, do nothing}
Normalize $\bm{\gamma} = \bm{\pi}/\sum\bm{\pi}$\;
Re-scale $\bm{\gamma} = \bm{\gamma}\cdot C$ \tcp*{keeping step sizes' magnitude}
\end{algorithm}

\begin{algorithm}[t]
\caption{p-Meta}\label{alg:micromaml}
\KwIn{meta-training task distribution $p(\mathsf{T})$, backbone $F$ with initial weights $\bm{w}$, meta attention parameters $\bm{w}^{\mathrm{atten}}$, inner step sizes $\bm{\alpha}$, outer step sizes $\beta$} 
\KwOut{meta-trained weights $\bm{w}$, meta-trained meta attention parameters $\bm{w}^{\mathrm{atten}}$, meta-trained sparse inner step sizes $\bm{\alpha}$}
\While {not done} {
    Sample a batch of $I$ tasks $\mathsf{T}^i\sim p(\mathsf{T})$\;
    \For {$i \leftarrow 1$ \KwTo $I$} {
        Update $\bm{w}^i$ in $K$ gradient descent steps with \eqref{eq:inner_atten}\;
    }
    Update $\bm{w}$ with \eqref{eq:outer_alpha}\;
    Update inner step sizes $\bm{\alpha}$ with \eqref{eq:regularization}\; 
    Update attention parameters $\bm{w}^{\mathrm{atten}}$ with \equref{eq:outer_atten}\;
}
\end{algorithm}

\fakeparagraph{Meta Attention Module Design}
\figref{fig:metaattention} (upper part) shows an example meta attention module.
We adapt the inference attention modules used in \cite{bib:CVPR18:Hu, bib:CVPR20:Chen}, yet with the following modifications.
\begin{itemize}
    \item 
    Unlike inference attention that applies to a single sample, training may calculate the averaged loss gradients based on a batch of samples.
    Since $\bm{g}(\bm{w}_l)$ does not have a batch dimension, the input to softmax function is first averaged over the batch data, see in \figref{fig:metaattention}.
    \item
    We enforce sparsity on the meta attention scores such that they can be utilized to save memory and computation in few-shot learning.
    The original attention in \cite{bib:CVPR18:Hu, bib:CVPR20:Chen} outputs normalized scales in $[0,1]$ from softmax.
    We clip the output with a clip ratio $\rho\in[0,1]$ to create zeros in $\bm{\gamma}$.
    This way, our meta attention modules yield batch-averaged sparse attention scores $\bm{\gamma}^{\mathrm{fw}}_{l}$ and $\bm{\gamma}^{\mathrm{bw}}_{l}$. 
    \algoref{alg:clip} shows this clipping and re-normalization process.
    Note that \algoref{alg:clip} is not differentiable.
    Hence we use the straight-through-estimator for its backward propagation in meta training. 
\end{itemize}

\begin{figure}[t]
\vspace{-0.5em}
  \centering
  \includegraphics[width=0.5\textwidth]{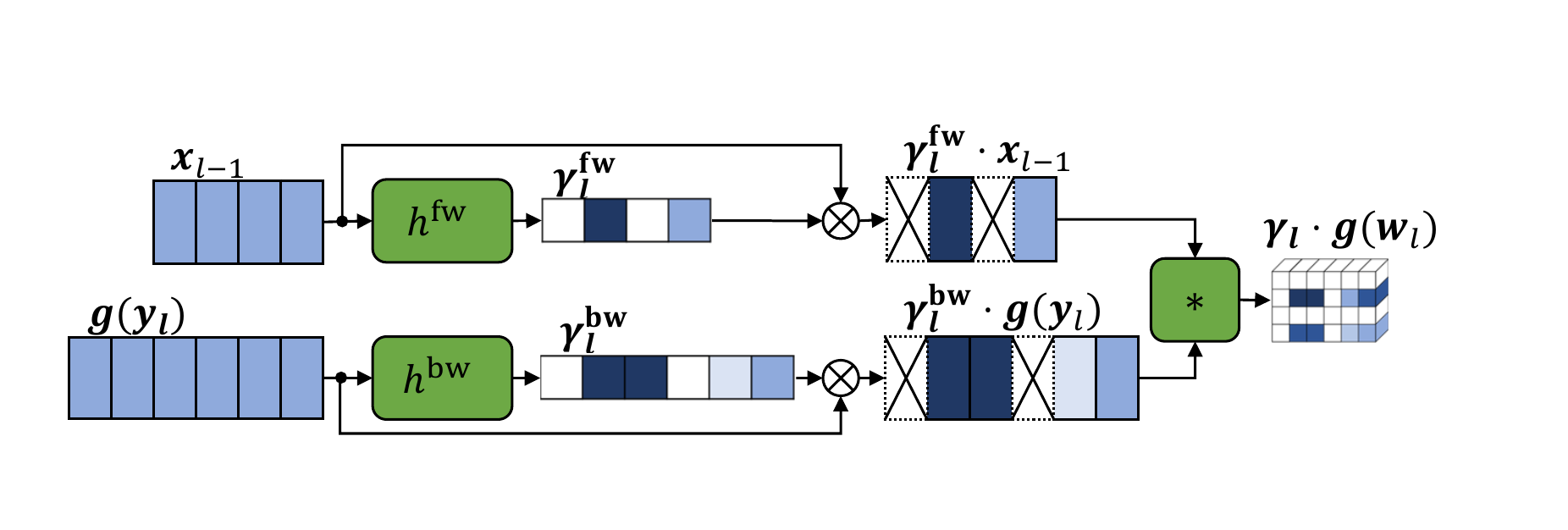} 
  \vspace{-2em}
  \caption{Meta attention of layer $l$ during on-device few-shot learning. Note that ``Forward'' part and ``Backward'' part are the same as \figref{fig:metaattention}, which are omitted for simplicity. Meta attention modules are not optimized during few-shot learning, thus are expressed as parameter-free functions $h^{\mathrm{fw}}$ and $h^{\mathrm{bw}}$. The input $\bm{x}_{l-1}$ stored during forward path is a sparse re-weighted tensor. }
\label{fig:metaattention_fewshot}
\end{figure}

\subsubsection{Exploiting Meta Attention for On-device Adaptation}
\label{sec:attention:fsl}
We now explain how to apply the meta attention to save memory during on-device few-shot learning.
Note that the parameters in the meta attention modules are fixed during few-shot learning.
Assume that at step $k$, layer $l$ has a nonzero step size $\alpha_l^k$.
In the forward pass, we only store a sparse tensor $\bm{\gamma}_l^{\mathrm{fw}}\cdot\bm{x}_{l-1}$, \ie its channels are stored only if they correspond to nonzero entries in $\bm{\gamma}_l^{\mathrm{fw}}$.
This reduces memory consumption as shown in \equref{eq:memorySimple} in Appendix \ref{app:analysis}.
Similarly, in the backward pass, we get a channel-wise sparse tensor $\bm{\gamma}_l^{\mathrm{bw}}\cdot\bm{g}(\bm{y}_{l})$.
Since both sparse tensors are used to calculate the corresponding nonzero gradients in $\bm{g}(\bm{w}_l)$, the computation cost is also reduced, see \equref{eq:MAC} in Appendix \ref{app:analysis}.
We plot the meta attention during on-device adaptation in \figref{fig:metaattention_fewshot}.

\subsection{Summary of \sysname}

\algoref{alg:micromaml} shows the overall process of \sysname during meta-training. 
The final meta-trained weights $\bm{w}$ from \algoref{alg:micromaml} are assigned to $\bm{w}^{\mathrm{meta}}$, see \secref{sec:background}.
The meta-trained backbone model $F(\bm{w}^{\mathrm{meta}})$, the sparse inner step sizes $\bm{\alpha}$, and the meta attention modules will be then deployed on edge devices and used to conduct a memory-efficient few-shot learning.

\subsection{Deployment Optimization}
\label{sec:others}
To further reduce the memory during few-shot learning, we propose gradient accumulation during backpropagation and replace batch normalization in the backbone with group normalization.

\subsubsection{Gradient Accumulation}
\label{sec:gradacc}
In standard few-shot learning, all the new samples (\eg $25$ for $5$-way $5$-shot) are fed into the model as one batch.
To reduce the peak memory due to large batch sizes, we conduct few-shot learning with gradient accumulation (GA).

GA is a technique that (\textit{i}) breaks a large batch into smaller partial batches; (\textit{ii}) sequentially forward/backward propagates each partial batches through the model; (\textit{iii}) accumulates the loss gradients of each partial batch and get the final averaged gradients of the full batch. 
Note that GA does not increase computation, which is desired for low-resource platforms.
We evaluate the impact of different sample batch sizes in GA in Appendix~\ref{app:batchsize}.

\subsubsection{Group Normalization}
\label{sec:norm}
Mainstream backbones in meta learning typically adopt batch normalization layers.
Batch normalization layers compute the statistical information in each batch, which is dependent on the sample batch size.
When using GA with different sample batch sizes, the inaccurate batch statistics can degrade the training performance (see Appendix \ref{app:pool}).
As a remedy, we use group normalization \cite{bib:ECCV18:Wu}, which does not rely on batch statistics (\ie independent of the sample batch size).
We also apply meta attention on group normalization layers when updating their weights.
The only difference w.r.t. \texttt{conv} and \texttt{fc} layers is that the stored input tensor (also the one used for the meta attention) is not $\bm{x}_{l-1}$, but its normalized version.

%% file: body/evaluation.tex
\section{Evaluation}
\label{sec:evaluation}
This section presents the evaluations of \sysname on standard few-shot image classification and reinforcement learning benchmarks.

\begin{table*}[t]
    \centering
    \caption{Few-shot image classification results on 4Conv and ResNet12. All methods are meta-trained on MiniImageNet, and are few-shot learned on the reported datasets: MiniImageNet, TieredImageNet, and CUB (denoted by Mini, Tiered, and CUB in the table). The total computation (\# GMACs) and the peak memory (MB) during few-shot learning are reported based on the theoretical analysis in Appendix~\ref{app:analysis}. }
    \label{tab:image}
    \footnotesize
    \begin{tabular}{llcccccccccc}
        \toprule
        \multicolumn{2}{l}{\multirow{2}{*}{Benchmarks}}                                 & \multicolumn{5}{c}{5-way 1-shot}                                                      & \multicolumn{5}{c}{5-way 5-shot}                                                      \\
                                                                                        \cmidrule(lr){3-7}                                                                      \cmidrule(lr){8-12}
                                    &                                                   & Mini   & Tiered & CUB                                 & Mini      & Mini              & Mini   & Tiered & CUB                                 & Mini      & Mini              \\
                                    &                                                   & \multicolumn{3}{c}{Accuracy}                          & GMAC      & Memory            & \multicolumn{3}{c}{Accuracy}                          & GMAC      & Memory            \\
        \midrule
        \multirow{6}{*}{4Conv}      & MAML \cite{bib:ICML17:Finn}                       & 46.2\% & 51.4\% & 39.7\%                              & 0.39      & 2.06              & 61.4\% & 66.5\% & 55.6\%                              & 1.96      & 2.06              \\
                                    & ANIL \cite{bib:ICLR20:Raghu}                      & 46.4\% & 51.5\% & 39.2\%                              & 0.14      & 0.92              & 60.6\% & 64.5\% & 54.2\%                              & 0.72      & 0.92              \\
                                    & BOIL \cite{bib:ICLR21:Oh}                         & 44.7\% & 51.3\% & 42.3\%                              & 0.39      & 2.05              & 60.5\% & 65.3\% & 58.3\%                              & 1.96      & 2.05              \\
                                    & MAML++ \cite{bib:ICLR19:Antreas}                  & 48.2\% & 53.2\% & \textbf{43.2\%}                     & 0.39      & 2.06              & 63.7\% & \textbf{68.5\%} & 59.1\%                     & 1.96      & 2.06              \\
                                    & \sysname (\ref{sec:LR})                           & 47.1\% & 52.3\% & 41.8\%                              & 0.16      & 1.00              & 62.9\% & 68.3\% & 59.3\%                              & 1.34      & 1.09              \\
                                    & \sysname (\ref{sec:LR}+\ref{sec:attention})       & \textbf{48.8\%} & \textbf{53.9\%} & 42.6\%            & 0.15      & 0.99              & \textbf{65.0\%} & \textbf{68.5\%} & \textbf{60.2\%}   & 1.11      & 1.04              \\
        \midrule
        \multirow{6}{*}{ResNet12}   & MAML \cite{bib:ICML17:Finn}                       & 51.7\% & 57.4\% & 41.3\%                              & 37.08     & 54.69             & 64.7\% & 69.6\% & 53.8\%                              & 185.42    & 54.69             \\
                                    & ANIL \cite{bib:ICLR20:Raghu}                      & 50.3\% & 56.7\% & 40.6\%                              & 12.42     & 3.62              & 62.3\% & 68.7\% & 54.0\%                              & 62.08     & 3.62              \\
                                    & BOIL \cite{bib:ICLR21:Oh}                         & 42.7\% & 47.7\% & 44.2\%                              & 37.08     & 54.69             & 53.6\% & 59.8\% & 53.7\%                              & 185.42    & 54.69             \\
                                    & MAML++ \cite{bib:ICLR19:Antreas}                  & 53.1\% & 58.6\% & 45.1\%                              & 37.08     & 54.69             & 68.6\% & \textbf{73.4\%} & 63.9\%                     & 185.42    & 54.69             \\
                                    & \sysname (\ref{sec:LR})                           & 51.8\% & 58.3\% & 40.6\%                              & 25.84     & 17.66             & 68.8\% & 72.6\% & 65.9\%                              & 124.15    & 18.95             \\
                                    & \sysname (\ref{sec:LR}+\ref{sec:attention})       & \textbf{53.6\%} & \textbf{59.4\%} & \textbf{45.4\%}   & 24.02     & 16.01             & \textbf{69.7\%} & 73.3\% & \textbf{66.6\%}            & 116.79    & 17.17             \\
    \bottomrule
    \end{tabular}
\end{table*}

\begin{table}[t]
    \centering
    \caption{Few-shot reinforcement learning results on 2D navigation and robot locomotion (larger return means better). A MLP with two hidden layers of size 100 is used as the policy model. The total computation (\# GMACs) and the peak memory (MB) during few-shot learning are reported based on the theoretical analysis in Appendix~\ref{app:analysis}.}
    \label{tab:mlp}
    \footnotesize
    \begin{tabular}{lcccccc}
        \toprule
        \multirow{2}{*}{Benchmarks}                     & \multicolumn{3}{c}{20 Rollouts}               & \multicolumn{3}{c}{20 Rollouts}           \\
                                                        \cmidrule(lr){2-4}                              \cmidrule(lr){5-7}                             
                                                        & \multicolumn{3}{c}{Half-Cheetah Velocity}     & \multicolumn{3}{c}{2D Navigation}         \\
                                            
                                                        & Return            & GMAC  & Memory            & Return            & GMAC  & Memory        \\
        \midrule
        MAML \cite{bib:ICML17:Finn}                     & -82.2             & 0.15  & 0.24              & -13.3             & 0.12  & 0.21          \\
        ANIL \cite{bib:ICLR20:Raghu}                    & -78.8             & 0.06  & 0.09              & -13.8             & 0.04  & 0.08          \\
        BOIL \cite{bib:ICLR21:Oh}                       & -76.4             & 0.15  & 0.23              & -12.4             & 0.12  & 0.21          \\
        MAML++ \cite{bib:ICLR19:Antreas}                & -69.6             & 0.15  & 0.24              & -17.6             & 0.12  & 0.21          \\
        p-Meta (\ref{sec:LR})                           & -65.5             & 0.11  & 0.12              & \textbf{-11.2}    & 0.09  & 0.09          \\
        p-Meta (\ref{sec:LR}+\ref{sec:attention})       & \textbf{-64.0}    & 0.11  & 0.11              & -11.8             & 0.09  & 0.09          \\
    \bottomrule
    \end{tabular}
\end{table}

\subsection{General Experimental Settings}
\label{sec:settings}

\fakeparagraph{Compared Methods}
We test the meta learning algorithms below. 
\begin{itemize}
    \item MAML \cite{bib:ICML17:Finn}: the original model-agnostic meta learning.
    \item ANIL \cite{bib:ICLR20:Raghu}: update the last layer only in few-shot learning.
    \item BOIL \cite{bib:ICLR21:Oh}: update the body except the last layer.
    \item MAML++ \cite{bib:ICLR19:Antreas}: learn a per-step per-layer step sizes $\bm{\alpha}$.
    \item \sysname (\ref{sec:LR}): can be regarded as a sparse version of MAML++, since it learns a sparse $\bm{\alpha}$ with our methods in \secref{sec:LR}. 
    \item \sysname (\ref{sec:LR}+\ref{sec:attention}): the full version of our methods which include the meta attention modules in \secref{sec:attention}. 
\end{itemize}
For fair comparison, all the algorithms are re-implemented with the deployment optimization in \secref{sec:others}. 

\fakeparagraph{Implementation}
The experiments are conducted with tools provided by TorchMeta \cite{bib:torchmeta, bib:torchmetarl}.
Particularly, the backbone is meta-trained with full sample batch size (\eg 25 for 5-way 5-shot) on meta training dataset.
After each meta training epoch, the model is tested (\ie few-shot learned) on meta validation dataset.
The model with the highest validation performance is used to report the final few-shot learning results on meta test dataset.
We follow the same process as TorchMeta \cite{bib:torchmeta, bib:torchmetarl} to build the dataset. 
During few-shot learning, we adopt a sample batch size of 1 to verify the model performance under the most strict memory constraints.

In \sysname, meta attention is applied to all \texttt{conv}, \texttt{fc}, and group normalization layers, except the last output layer, because (\textit{i}) we find modifying the last layer's gradients may decrease accuracy; (\textit{ii}) the final output is often rather small in size, resulting in little memory saving even if imposing sparsity on the last layer.  
Without further notations, we set $\rho=0.3$ in forward attention, and $\rho=0$ in backward attention across all layers, as the sparsity of $\bm{\gamma}_l^{\mathrm{bw}}$ almost has no effect on the memory saving.

\fakeparagraph{Metrics}
We compare the peak memory and MACs of different algorithms.
Note that the reported peak memory and MACs for \sysname also include the consumption from meta attention, although they are rather small related to the backward propagation.

\subsection{Performance on Image Classification}
\label{sec:image}

\fakeparagraph{Settings}
We test on standard few-shot image classification tasks (both in-domain and cross-domain). 
We adopt two common backbones, ``4Conv'' \cite{bib:ICML17:Finn} which has 4 \texttt{conv} blocks with 32 channels in each block, and ``ResNet12'' \cite{bib:NIPS18:Oreshkin} which contains 4 residual blocks with $\{64,128,256,512\}$ channels in each block respectively.
We replace the batch normalization layers with group normalization layers, as discussed in \secref{sec:norm}.
We experiment in both 5-way 1-shot and 5-way 5-shot settings.
We train the model on MiniImageNet \cite{bib:NIPS16:Vinyals} (both meta training and meta validation dataset) with 100 meta epochs.
In each meta epoch, 1000 random tasks are drawn from the task distribution. 
The task batch size is set to 4 in general, except for ResNet12 under 5-way 5-shot settings where we use 2.
The model is updated with 5 gradient steps (\ie $K=5$) in both inner loop of meta-training and few-shot learning.  
We use Adam optimizer with cosine learning rate scheduling as \cite{bib:ICLR19:Antreas} for all outer loop updating.
The (initial) inner step size $\bm{\alpha}$ is set to 0.01.
The meta-trained model is then tested on three datasets MiniImageNet \cite{bib:NIPS16:Vinyals}, TieredImageNet \cite{bib:ICLR18:Ren}, and CUB \cite{bib:CUB} to verify both \textit{in-domain} and \textit{cross-domain} performance.

\fakeparagraph{Results}
\tabref{tab:image} shows the accuracy averaged over $5000$ new unseen tasks randomly drawn from the meta test dataset.
We also report the average number of GMACs and the average peak memory per task according to Appendix~\ref{app:analysis}.
Clearly, \sysname almost always yields the highest accuracy in all settings. 
Note that the comparison between ``\sysname (\ref{sec:LR})'' and ``MAML++'' can be considered as the ablation studies on learning sparse layer-wise inner step sizes proposed in \secref{sec:LR}. 
Thanks to the imposed sparsity on $\bm{\alpha}$, ``\sysname (\ref{sec:LR})'' significantly reduces the peak memory ($2.5\times$ saving on average and up to $3.1\times$) and the computation burden ($1.7\times$ saving on average and up to $2.4\times$) over ``MAML++''. 
Note that the imposed sparsity also cause a moderate accuracy drop.
However, with the meta attention, ``\sysname (\ref{sec:LR}+\ref{sec:attention})'' not only notably improves the accuracy but also further reduces the peak memory ($2.7\times$ saving on average and up to $3.4\times$) and computation ($1.9\times$ saving on average and up to $2.6\times$) over ``MAML++''.  
``ANIL'' only updates the last layer, and therefore consumes less memory but also yields a substantially lower accuracy.

\subsection{Performance on Reinforcement Learning}
\label{sec:reinforcement}

\fakeparagraph{Settings}
To show the versatility of \sysname, we experiment with two few-shot reinforcement learning problems: 2D navigation and Half-Cheetah robot locomotion simulated with MuJoCo library \cite{bib:IROS12:Todorov}.
For all experiments, we mainly adopt the experimental setup in \cite{bib:ICML17:Finn}.
We use a neural network policy with two hidden \texttt{fc} layers of size 100 and ReLU activation function.
We adopt vanilla policy gradient \cite{bib:ML1992:Williams} for the inner loop and trust-region policy optimization \cite{bib:ICML15:Schulman} for the outer loop.
During the inner loop as well as few-shot learning, the agents rollout 20 episodes with a horizon size of 200 and are updated for one gradient step.
The policy model is trained for 500 meta epochs, and the model with the best average return during training is used for evaluation.
The task batch size is set to 20 for 2D navigation, and 40 for robot locomotion.
The (initial) inner step size $\bm{\alpha}$ is set to 0.1.
Each episode is considered as a data sample, and thus the gradients are accumulated 20 times for a gradient step.

\fakeparagraph{Results}
\tabref{tab:mlp} lists the average return averaged over 400 new unseen tasks randomly drawn from simulated environments.
We also report the average number of GMACs and the average peak memory per task according to Appendix \ref{app:analysis}.
Note that the reported computation and peak memory do not include the estimations of the advantage \cite{bib:ICML16:Duan}, as they are relatively small and could be done during the rollout. 
\sysname consumes a rather small amount of memory and computation, while often obtains the highest return in comparison to others. 
Therefore, \sysname can fast adapt its policy to reach the new goal in the environment with less on-device resource demand.

\subsection{Ablation Studies on Meta Attention}
\label{sec:ablation}
We study the effectiveness of our meta attention via the following two ablation studies.
The experiments are conducted on ``4Conv'' in both 5-way 1-shot and 5-way 5-shot as \secref{sec:image}.

\fakeparagraph{Sparsity in Meta Attention}
\tabref{tab:ablation} shows the few-shot classification accuracy with different sparsity settings in the meta attention. 

We first do not impose sparsity on $\bm{\gamma}_l^{\mathrm{fw}}$ and $\bm{\gamma}_l^{\mathrm{bw}}$ (\ie set both $\rho$'s as 0), and adopt forward attention and backward attention separately. 
In comparison to no meta attention at all, enabling either forward or backward attention improves accuracy. 
With both attention enabled, the model achieves the best performance.

We then test the effects when imposing sparsity on $\bm{\gamma}_l^{\mathrm{fw}}$ or $\bm{\gamma}_l^{\mathrm{bw}}$ (\ie set $\rho>0$). 
We use the same $\rho$ for all layers.
We observe a sparse $\bm{\gamma}_l^{\mathrm{bw}}$ often cause a larger accuracy drop than a sparse $\bm{\gamma}_l^{\mathrm{fw}}$.
Since a sparse $\bm{\gamma}_l^{\mathrm{bw}}$ does not bring substantial memory or computation saving (see Appendix~\ref{app:analysis}), we use $\rho=0$ for backward attention and $\rho=0.3$ for forward attention.

Attention scores $\bm{\gamma}_l$ introduce a dynamic channel-wise learning rate according to the new data samples. 
We further compare meta attention with a static channel-wise learning rate, where the channel-wise learning rate $\bm{\alpha}^{\mathrm{Ch}}$ is meta-trained as the layer-wise inner step sizes in \secref{sec:LR} while without imposing sparsity.
By comparing ``$\bm{\alpha}^{\mathrm{Ch}}$'' with ``0, 0'' in \tabref{tab:ablation}, we conclude that the dynamic channel-wise learning rate yields a significantly higher accuracy.

\begin{table}[t]
    \centering
    \caption{Ablation results of meta attention on 4Conv.}
    \label{tab:ablation}
    \footnotesize
    \begin{tabular}{cccccccc}
        \toprule
        \multicolumn{2}{c}{$\rho$}                          & \multicolumn{3}{c}{5-way 1-shot}                      & \multicolumn{3}{c}{5-way 5-shot}              \\
        \cmidrule(lr){1-2}                                  \cmidrule(lr){3-5}                                      \cmidrule(lr){6-8}
        fw      & bw                                        & Mini   & Tiered & CUB                                 & Mini   & Tiered & CUB                         \\
        \midrule
        x       & x                                         & 47.1\% & 52.3\% & 41.8\%                              & 62.9\% & 68.3\% & 59.3\%                      \\
        0       & x                                         & 48.1\% & 53.2\% & 41.7\%                              & 64.1\% & 68.4\% & 59.0\%                      \\
        x       & 0                                         & 47.8\% & 53.1\% & 40.9\%                              & 63.9\% & 68.5\% & 60.0\%                      \\
        0       & 0                                         & \textbf{49.0\%} & \textbf{54.2\%} & \textbf{43.1\%}   & 64.5\% & \textbf{69.2\%} & \textbf{60.2\%}    \\
        0       & 0.3                                       & 48.5\% & 53.4\% & 42.2\%                              & 64.7\% & 68.2\% & 59.3\%                      \\
        0.3     & 0                                         & 48.8\% & 53.9\% & 42.6\%                              & \textbf{65.0\%} & 68.5\% & \textbf{60.2\%}    \\
        0.3     & 0.3                                       & 48.7\% & 53.7\% & 42.3\%                              & 64.5\% & 68.3\% & 59.5\%                      \\
        0.5     & 0.5                                       & 48.2\% & 53.4\% & 42.7\%                              & 64.8\% & 68.1\% & 59.1\%                      \\
        \multicolumn{2}{c}{$\bm{\alpha}^{\mathrm{Ch}}$}     & 47.8\% & 52.8\% & 41.0\%                              & 63.6\% & 68.1\% & 58.1\%                      \\
    \bottomrule
    \end{tabular}
    \begin{tablenotes}
        \footnotesize
        \item
        x: no forward/backward meta attention, \ie $\bm{\gamma}_l^{\mathrm{fw}}=1$ or $\bm{\gamma}_l^{\mathrm{bw}}=1$.
        \item
        $\bm{\alpha}^{\mathrm{Ch}}$: introducing an input- and output-channel-wise inner step sizes $\bm{\alpha}^{\mathrm{Ch}}$ per layer. We use $\bm{\alpha}\cdot\bm{\alpha}^{\mathrm{Ch}}$ as the overall inner step sizes. $\bm{\alpha}^{\mathrm{Ch}}$ is meta-trained as $\bm{\alpha}$ without imposing sparsity.
    \end{tablenotes}
\end{table}

\begin{figure*}[t]
\vspace{-0.5em}
  \centering
  \includegraphics[width=0.99\textwidth]{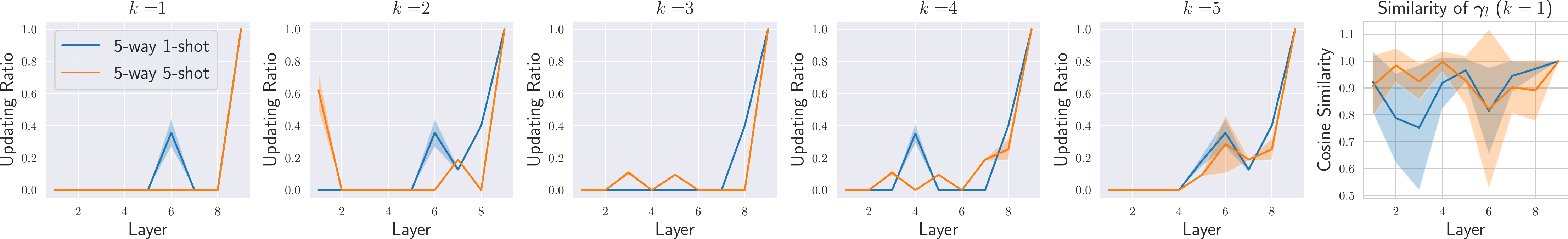} 
  \vspace{-0.5em}
  \caption{\textit{Left (1:5):} Layer-wise updating ratios (mean $\pm$ standard deviation) in each updating step. Note that the ratio of updated weights is determined by both static layer-wise inner step sizes $\alpha_{1:L}^{1:K}$ and the dynamic meta attention scores $\bm{\gamma}_{1:L}$. The layer with an updating ratio of 0 means its $\alpha=0$. \textit{Right:} Cosine similarity (mean $\pm$ standard deviation) of $\bm{\gamma}_{1:L}$ between random pair of data samples. The results are reported in step 1, because all samples are fed into the same initial model in step 1. }
\label{fig:ratio}
\end{figure*}

\fakeparagraph{Layer-wise Updating Ratios}
To study the resulted updating ratios across layers, \ie the layer-wise sparsity of weight gradients, we randomly select 100 new tasks and plot the layer-wise updating ratios, see \figref{fig:ratio} \textit{Left (1:5)}. 
The ``4Conv'' backbone has 9 layers ($L=9$), \ie 8 alternates of \texttt{conv} and group normalization layers, and an \texttt{fc} output layer.
As mentioned in \secref{sec:settings}, we do not apply meta attention to the output layer, \ie $\bm{\gamma}_9=1$.
The used backbone is updated with 5 gradient steps ($K=5$). 
We use $\rho=0.3$ for forward attention, and $\rho=0$ for backward. 
Note that \algoref{alg:clip} adaptively determines the sparsity of $\bm{\gamma}_l$, which also means different samples may result in different updating ratios even with the same $\rho$ (see \figref{fig:ratio}). 
The size of $\bm{x}_{l-1}$ often decreases along the layers in current DNNs.
As expected, the latter layers are preferred to be updated more, since they need a smaller amount of memory for updating.
Interestingly, even if with a small $\rho$($=0.3$), the ratio of updated weights is rather small, \eg smaller than $0.2$ in step 3 of 5-way 5-shot.
It implies that the outputs of softmax have a large discrepancy, \ie only a few channels are adaptation-critical for each sample, which in turn verifies the effectiveness of our meta attention mechanism. 

We also randomly pair data samples and compute the cosine similarity between their attention scores $\bm{\gamma}_l$.  
We plot the cosine similarity of step 1 in \figref{fig:ratio} \textit{Right}. 
The results show that there may exist a considerable variation on the adaptation-critical weights selected by different samples, which is consistent with our observation in \tabref{tab:ablation}, \ie dynamic learning rate outperforms the static one.

%% file: body/related.tex
\section{Related Work}
\label{sec:related}

\fakeparagraph{Meta Learning for Few-Shot Learning}
Meta learning is a prevailing solution to few-shot learning \cite{bib:arXiv20:Hospedales}, where the meta-trained model can learn an unseen task from a few training samples, \ie data-efficient adaptation.
The majority of meta learning methods can be divided into two categories, (\textit{i}) metric-based methods \cite{bib:NIPS16:Vinyals, bib:NIPS17:Snell, bib:CVPR18:Sung} that learn an embedded metric for classification tasks to map the query samples onto the classes of labeled support samples, (\textit{ii}) gradient-based methods \cite{bib:ICLR19:Antreas, bib:ICML17:Finn, bib:ICLR20:Raghu, bib:ICLR20:Triantafillou, bib:ICLR21:Oh, bib:NIPS21:Oswald} that learn an initial model (and/or optimizer parameters) such that it can be adapted with gradient information calculated on the new few samples.   
In comparison to metric-based methods, we focus on gradient-based meta learning methods for their wide applicability in various learning tasks (\eg regression, classification, reinforcement learning) and the availability of gradient-based training frameworks for low-resource devices \cite{bib:tfliteOndeviceTraining}.

Particularly, we aim at meta training a DNN that allows effective adaptation on memory-constrained devices.
Most meta learning algorithms \cite{bib:ICLR19:Antreas, bib:ICML17:Finn, bib:NIPS21:Oswald} optimize the backbone network for better generalization yet ignore the workload if the meta-trained backbone is deployed to low-resource platforms for model adaptation. 
Manually fixing certain layers during on-device few-shot learning \cite{bib:ICLR20:Raghu, bib:ICLR21:Oh, bib:AAAI21:Shen} may also reduce memory and computation, but to a much lesser extent as shown in our evaluations.

\fakeparagraph{Efficient DNN Training}
Existing efficient training schemes are mainly designed for high-throughput GPU training on large-scale datasets.
A general strategy is to trade memory with computation \cite{bib:arXiv16:Chen, bib:NIPS16:Gruslys}, which is unfit for IoT device with a limited computation capability. 
An alternative is to sparsify the computational graphs in backpropagation \cite{bib:NIPS20:Raihan}.
Yet it relies on massive training iterations on large-scale datasets.
Other techniques include layer-wise local training \cite{bib:NIPS17:Greff} and reversible residual module \cite{bib:NIPS17:Gomez}, but they often incur notable accuracy drops. 

There are a few studies on DNN training on low-resource platforms, such as updating the last several layers only \cite{bib:MLSys21:Mathur}, reducing batch sizes \cite{bib:SenSys19:Lee}, and gradient approximation \cite{bib:ICASSP20:Gooneratne}.
However, they are designed for vanilla supervised training, \ie train and test on the same task. 
One recent study proposes to update the bias parameters only for memory-efficient transfer learning \cite{bib:NIPS20:Cai}, yet transfer learning is prone to overfitting when trained with limited data \cite{bib:ICML17:Finn}.

%% file: body/conclusion.tex
\section{Conclusion}
\label{sec:conclusion}
In this paper, we present \sysname, a new meta learning scheme for data- and memory-efficient on-device DNN adaptation.
It enables structured partial parameter updates for memory-efficient few-shot learning by automatically identifying adaptation-critical weights both layer-wise and channel-wise. 
Evaluations show a reduction in peak dynamic memory by 2.5$\times$ on average over the state-of-the-art few-shot adaptation methods.
We envision \sysname as an early step towards adaptive and autonomous edge intelligence applications.

\section*{Acknowledgement}
Part of Zhongnan Qu and Lothar Thiele's work was supported by the Swiss National Science Foundation in the context of the NCCR Automation.
This research was supported by the Lee Kong Chian Fellowship awarded to Zimu Zhou by Singapore Management University.
Zimu Zhou is the corresponding author.

%% file: body/appendix.tex
\appendix

\section{Memory and Computation}
\label{app:analysis}
In the following, we derive the memory requirement and computation workload for inference and adaptation. We restrict ourselves to a feed-forward network of fully-connected (\texttt{fc}) or convolutional (\texttt{conv}) layers. 
Note that our analysis focuses on 2D \texttt{conv} layers but can apply to other \texttt{conv} layer types as well. 
We assume the rectified linear activation function (ReLU) for all layers, denoted as $\sigma$. For simplicity, we omit the bias, normalization layers, pooling or strides. 
We use the notation $m(\bm{x})$ to denote the memory demand in words to store tensor $\bm{x}$. 
The wordlength is denoted as $\mathit{T}$. 

For representing indexed summations we use the Einstein notation. 
If index variables appear in a term on the right hand side of an equation and are not otherwise defined (free indices), it implies summation of that term over the range of the free indices. 
If indices of involved tensor elements are out of range, the values of these elements are assumed to be 0.

\subsection{Single Layer}

We start with a single layer and accumulate the memory and computation for networks with several layers afterwards. 
Assume the input tensor of a layer is $\bm{x}$, the weight tensor is $\bm{w}$, the result after the linear transformation is $\bm{y}$, and the layer output after the non-linear operator is $\bm{z}$ which is also the input to the next layer. 

For convolutional layers, we have $\bm{x} \in \mathbb{R}^{C_I \times H_I \times W_I}$ and elements $x_{cij}$, where $C_I$, $H_I$, and $W_I$ denote the number of input channels, height and width, respectively. 
In a similar way, we have $\bm{z} \in \mathbb{R}^{C_O \times H_O \times W_O}$ with elements $x_{fij}$ where $C_O$, $H_O$, and $W_O$ denote the number of output channels, height and width, respectively. Moreover, $\bm{w} \in \mathbb{R}^{C_O \times C_I \times S \times S}$ with elements $w_{fcmn}$. 
Therefore,
\[
\medop{m(\bm{x})= C_I H_I W_I \; , \; \; m(\bm{y})= m(\bm{z}) = C_O H_O W_O \; , \; \; m(\bm{w}) = C_O C_I S^2}
\]
For fully connected layers we have $\bm{x} \in \mathbb{R}^{C_I}$, $\bm{y}, \bm{z} \in \mathbb{R}^{C_O}$, and $\bm{w} \in \mathbb{R}^{C_O \times C_I}$ with memory demand
\[
\medop{m(\bm{x})= C_I  \; , \; \; m(\bm{y})= m(\bm{z}) = C_O  \; , \; \; m(\bm{w}) = C_O C_I} 
\]

\subsubsection{Fully Connected Layer}
For inference we derive the relations $y_{f} = w_{fc} x_c$ and $z_{f} = \sigma(y_{f})$ for all admissible indices $f \in [1, C_O]$. 
The necessary dynamic memory has a size of about $m(\bm{x}) + m(\bm{y})$ words and we need about $m(\bm{w})$ MAC operations. 

For adaptation, we suppose that $\frac{\partial \ell}{\partial z_i}$ is already provided from the next layer. 
We find $\frac{\partial \ell}{\partial y_i} = \sigma'(y_i) \cdot \frac{\partial \ell}{\partial z_i}$ with $\sigma'(y_i) = \begin{cases} 1 & \mbox{if } y_i > 0 \\ 0 & \mbox{if } y_i < 0  \end{cases}$ which leads to $\frac{\partial \ell}{\partial x_i} = w_{ji} \cdot \frac{\partial \ell}{\partial y_j}$. 
The necessary dynamic memory is about $m(\bm{x}) + m(\bm{y}) \cdot (1 + \frac{1}{\mathit{T}})$ words, where the last term comes from storing $\sigma'(y_i)$ single bits from the forward path. 
We need about $m(\bm{w})$ MAC operations.

According to the approach described in the paper we are only interested in the partial derivatives $\frac{\partial \ell}{\partial w_{fc}}$ if $\alpha > 0$ for this layer, and if scales $\gamma^{\mathrm{bw}}_{f} > 0$ and $\gamma^{\mathrm{fw}}_{c} > 0$ for indices $f$, $c$. 
To simplify the notation, let us define the critical ratios 
\begin{gather*}
\medop{\mu^{\mathrm{fw}} = \frac{\text{number of nonzero elements of } \gamma^{\mathrm{fw}}_{c}}{C_I}} \\
\medop{\mu^{\mathrm{bw}} = \frac{\text{number of nonzero elements of } \gamma^{\mathrm{bw}}_{f}}{C_O}}
\end{gather*}
which are 1 if all channels are determined to be critical for weight adaptation, and 0 if none of them. 

We find $\gamma^{\mathrm{bw}}_{f} \frac{\partial \ell}{\partial w_{fc}}  \gamma^{\mathrm{fw}}_{c}  =  (\gamma^{\mathrm{bw}}_{f} \frac{\partial \ell}{\partial y_{f}}) \cdot  (\gamma^{\mathrm{fw}}_{c} x_c)$.
Therefore, we need $\mu^{\mathrm{fw}} \mu^{\mathrm{bw}} m(\bm{w}) + \mu^{\mathrm{fw}} m(\bm{x})$ words dynamic memory if $\alpha > 0$ where the latter term considers the information needed from the forward path. 
We require about $\mu^{\mathrm{fw}} \mu^{\mathrm{bw}} m(\bm{w})$ MAC operations if $\alpha > 0$.

\subsubsection{Convolutional Layer}

The memory analysis for a convolutional layer is very similar, just replacing matrix multiplication by convolution. 
For inference we find $y_{fij} = w_{fcmn} x_{c, i+m-1, j+n-1}$ and $z_{fij} = \sigma(y_{fij})$ for all admissible indices $f$, $i$, $j$. 
The necessary dynamic memory has a size of about $\max \{ m(\bm{x}), m(\bm{y}) \}$ words when using memory sharing between input and output tensors. 
We need about $H_O W_O \cdot m(\bm{w})$ MAC operations. 

For adaptation, we again suppose that $\frac{\partial \ell}{\partial z_{fij}}$ is provided from the next layer. 
We find $\frac{\partial \ell}{\partial y_{fij}} = \sigma'(y_{fij}) \cdot \frac{\partial \ell}{\partial z_{fij}}$ and get $\frac{\partial \ell}{\partial x_{cij}} = w_{fcmn}  \cdot \frac{\partial \ell}{\partial y_{f, i + m - 1, j + n - 1}}$. 
The necessary memory is about $\max \{ m(\bm{x}), m(\bm{y}) \} +  \frac{ m(\bm{y})}{\mathit{T}}$ words, where the last term comes from storing $\sigma'(y_{fij})$ single bits from the forward path. 
We need about $H_I W_I \cdot m(\bm{w})$ multiply and accumulate operations.

For determining the weight gradients we find $\frac{\partial \ell}{\partial w_{fcmn}}  =  \frac{\partial \ell}{\partial y_{fij}} \cdot x_{c, i + m -1, j + n - 1}$. 
When considering the scales for filtering, we yield $\gamma^{\mathrm{bw}}_{f} \frac{\partial \ell}{\partial w_{fcmn}}  \gamma^{\mathrm{fw}}_{c}  =  (\gamma^{\mathrm{bw}}_{f} \frac{\partial \ell}{\partial y_{fij}}) \cdot  (\gamma^{\mathrm{fw}}_{c} x_{c, i + m -1, j + n - 1})$. 
As a result, we need $\mu^{\mathrm{fw}} \mu^{\mathrm{bw}} m(\bm{w}) +  \mu^{\mathrm{fw}} m(\bm{x})$ words of dynamic memory if $\alpha > 0$ where the latter term considers the information needed from the forward path. 
We require about $\mu^{\mathrm{fw}} \mu^{\mathrm{bw}} H_O W_O m(\bm{w})$ MAC operations if $\alpha > 0$. 

Finally, let us determine the required memory and computation to determine the scales $\gamma^{\mathrm{fw}}_{c}$ and $\gamma^{\mathrm{bw}}_{f}$. 
According to \figref{fig:metaattention}, we find as an upper bound for the memory $B \cdot ( C_I + C_O)$ and $( C_I H_I W_I + 2 C_I^2 + C_O H_O W_O + 2 C_O^2 )$ MAC operations.

\subsection{All Layers}

The above relations are valid for a single layer. 
The following relations hold for the overall network. 
In order to simplify the notation, we consider a network that consists of convolution layers only.
Extensions to mixed layers can simply be done.

We suppose $L$ layers with sizes $C_l$, $H_l$, $W_l$ and $S_l$ for the number of output channels, output width, output height and kernel size, respectively. 
We assume that the step-sizes $\alpha_l$ for some iteration of the adaption are given. 
The memory requirement in words is
\[
\medop{m(\bm{x}_l)= C_l H_l W_l \; , \; \; m(\bm{w}_l) = C_l C_{l-1} S_l^2}
\]
and the word-length is again denoted as $\mathit{T}$.
We define as $\hat{\alpha}_l = \begin{cases} 1 & \mbox{if } \alpha_l > 0 \\ 0 & \mbox{if } \alpha_l = 0  \end{cases}$ the mask that determines whether the weight adaptation for this layer is necessary or not.

Let us first look at the forward path. 
The necessary dynamic memory is about $\max_{0 \leq l \leq L} \{ m(\bm{x}_l) \}$ words. 
The amount of MAC operations is $\sum_{1 \leq l \leq L} H_l W_l m(\bm{w}_l)$. 

The backward path needs only to be evaluated until we reach the first layer where we require the computation of the gradients. 
We define $l_\mathit{min} = \min \{ l \, | \, \hat{\alpha}_{l} = 1 \}$. 
For the calculation of the partial derivatives of the activations we need dynamic memory of $\max_{l_{\mathit{min}} \leq l \leq L} \{ m(\bm{x}_l) \} + \frac{1}{\mathit{T}} \sum_{l_\mathit{min} \leq l \leq L} m(\bm{x}_l)$ words where the last term is due to storing the derivatives of the ReLU operations.
We need about $\sum_{l_\mathit{min} + 1 \leq l \leq L} H_{l-1} W_{l-1} m(\bm{w}_l) $ MAC operations.

The second contribution of the backward path is for computing the weight gradients. 
The memory and computation demand of the scales will be neglected as they are much smaller than other contributions. 
We can determine the necessary dynamic memory as $\max_{1 \leq l \leq L} \{ \hat{\alpha}_l \mu^{\mathrm{fw}}_l \mu^{\mathrm{bw}}_l m(\bm{w}_l)\} + \sum_{1 \leq l \leq L} \hat{\alpha}_l \mu^{\mathrm{fw}}_l m(\bm{x}_{l-1})$, and we need $\sum_{1 \leq l \leq L}  \hat{\alpha}_l \mu^{\mathrm{fw}}_l \mu^{\mathrm{bw}}_l H_l W_l m(\bm{w}_l)$  MAC operations. 

Considering all necessary dynamic memory with memory reuse for an adaptation step, we get an estimation of memory in words
\begin{multline}
\label{eq:memoryAll}
\medop{\max_{0 \leq l \leq L} \{ m(\bm{x}_l) \}  + \sum_{1 \leq l \leq L} \hat{\alpha}_l m(\bm{w}_l) +} \\ \medop{+ \sum_{1 \leq l \leq L} \hat{\alpha}_l \mu^{\mathrm{fw}}_l m(\bm{x}_{l-1}) + \frac{1}{\mathit{T}} \sum_{l_\mathit{min} \leq l \leq L} m(\bm{x}_l)}
\end{multline}
if we accumulate the weight gradients before doing an SGD step and re-use some memory during back-propagation. More elaborate memory re-use can be used to slightly sharpen the bounds without a major improvement.
For conventional training, each parameter is in 32-bit floating point format, \ie one word corresponds to 32-bit. As discussed in \secref{sec:background}, we only consider max-pooling and ReLU-styled activation as the $\sigma$ function. The wordlength $T$ in \equref{eq:memoryAll} is set as 16 for max-pooling , and 32 for ReLU-styled activation.
One can see that under the typical assumptions for network parameters, the above memory requirement in words is dominated by 
\begin{equation}
\label{eq:memorySimple}
\medop{\sum_{1 \leq l \leq L} \hat{\alpha}_l \mu^{\mathrm{fw}}_l m(\bm{x}_{l-1})}
\end{equation}
The necessary storage between the forward and backward path is reduced proportionally to $\mu^{\mathrm{fw}}_l$ with factor $m(\bm{x}_{l-1})$. 

Finally, the amount of MAC computations can be estimated as 
\begin{equation}
\label{eq:MAC}
\medop{\sum_{1 \leq l \leq L} H_l W_l m(\bm{w}_l) ( 1 + 
\hat{\alpha}_l \mu^{\mathrm{fw}}_l \mu^{\mathrm{bw}}_l) + 
\sum_{l_\mathit{min} \leq l \leq L} H_{l-1} W_{l-1} m(\bm{w}_l)}
\end{equation}
while neglecting lower order terms. Here it is important to note that all terms are of similar order. The approach used in the paper does not determine a trade-off between computation and memory, but reduces the amount of MAC operations. This reduction is less than the reduction in required dynamic memory.

\section{Other Experiments}
\label{app:other}

\subsection{Pooling \& Normalization}
\label{app:pool}

In this section, we test the backbone network with different types of pooling and normalization. 
Without further notations in the following experiments, we meta-train our ``4Conv'' backbone on MiniImageNet with full batch sizes, and conduct few-shot learning with gradient accumulation with a batch size of 1, as in \secref{sec:evaluation}.
Here, we report the results with the original ``MAML'' method \cite{bib:ICML17:Finn} in \tabref{tab:pool}. 
Clearly, the discrepancy of batch statistics between meta-training phase and few-shot learning phase causes a large accuracy loss in batch normalization layers.
Batch normalization works only if few-shot learning uses full batch sizes, \ie without gradient accumulation, which however does not fit in our memory-constrained scenarios (see \secref{sec:gradacc}).
In addition, max-pooling performs better than average-pooling.
We thus use group normalization and max-pooling in our backbone model, see \secref{sec:evaluation}.

\begin{table}[t]
    \centering
    \footnotesize
    \caption{Comparison between different pooling and normalization layers.}
    \label{tab:pool}
    \begin{tabular}{ccccc}
        \toprule
        \multicolumn{2}{c}{4Conv}       & \multicolumn{3}{c}{5-way 1-shot}      \\
        \cmidrule(lr){1-2}              \cmidrule(lr){3-5}                              
        Pooling     & Norm.             & Mini   & Tiered & CUB                 \\
        \midrule
        Average     & Batch             & 25.3\% & 27.2\% & 26.1\%              \\
        Average     & Group             & 45.8\% & 50.3\% & 40.2\%              \\
        Max         & Batch             & 27.6\% & 28.9\% & 26.5\%              \\
        Max         & Group             & 46.2\% & 51.4\% & 39.9\%              \\
    \bottomrule
    \end{tabular}
\end{table}

\subsection{Sample Batch Size}
\label{app:batchsize}

In this section, we show the effects brought from different sample batch sizes.
During few-shot learning phase, gradient accumulation is applied to fit in different on-device memory constraints. 
We report the accuracy when adopting different sample batch sizes in gradient accumulation. 
Although group normalization eliminates the variance of batch statistics, adopting different batch sizes may still result in diverse performance due to the batch-averaged scores in meta attention.
The results in \tabref{tab:batchsize} show that different batch sizes yield a similar accuracy level, which indicates that our meta attention module is relatively robust to batch sizes.

\begin{table}[t]
    \centering
    \footnotesize
    \caption{Ablation results of sample batch sizes.}
    \label{tab:batchsize}
    \begin{tabular}{lcccccc}
        \toprule
                                    & \multicolumn{3}{c}{5-way 1-shot}              & \multicolumn{3}{c}{5-way 5-shot}              \\
                                    \cmidrule(lr){2-4}                              \cmidrule(lr){5-7}
        Batch Size                  & 1      & 2      & 5                           & 1      & 5      & 25                          \\
        \midrule
        Mini                        & 48.8\% & 48.7\% & 48.3\%                      & 65.0\% & 65.1\% & 64.7\%                      \\
        Tiered                      & 53.9\% & 53.6\% & 54.3\%                      & 68.5\% & 68.9\% & 68.1\%                      \\
        CUB                         & 42.6\% & 42.1\% & 42.4\%                      & 60.2\% & 59.5\% & 60.6\%                      \\
    \bottomrule
    \end{tabular}
\end{table}

\subsection{Sparse \textit{x} and Sparse \textit{g}(\textit{y})}
\label{app:sparse}

Our meta attention modules take $\bm{x}_{l-1}$ and $\bm{g}(\bm{y}_l)$ as inputs, and output attention scores which are used to create sparse $\bm{g}(\bm{w}_l)$.
However, applying the resulted sparse attention scores on $\bm{x}_{l-1}$ and $\bm{g}(\bm{y}_l)$ can also bring memory and computation benefits, as discussed in \secref{sec:overview}.
We conduct the ablations when multiplying attention scores $\bm{\gamma}_l^{\mathrm{fw}}$ and $\bm{\gamma}_l^{\mathrm{bw}}$ on $\bm{g}(\bm{w}_l)$ (also the one used in the main text), or on $\bm{x}_{l-1}$ and $\bm{g}(\bm{y}_{l})$ respectively.
The results in \tabref{tab:sparse} show that a channel-wise sparse $\bm{x}_{l-1}$ hugely degrades the performance, in comparison to only imposing sparsity on $\bm{g}(\bm{w}_l)$ while using a dense $\bm{x}_{l-1}$ in the forward pass.  
In addition, directly adopting a sparse $\bm{g}(\bm{y}_{l})$ in backpropagation may even cause non-convergence in few-shot learning. 
We think this is due to the fact that the error accumulates along the propagation when imposing sparsity on $\bm{x}_{l-1}$ or $\bm{g}(\bm{y}_{l})$.

\begin{table}[t]
    \centering
    \footnotesize
    \caption{Ablation results of sparse $\bm{x}_{l-1}$ and sparse $\bm{g}(\bm{y}_l)$.}
    \label{tab:sparse}
    \begin{tabular}{cccccccc}
        \toprule
        \multicolumn{2}{c}{$\rho=0.3$}                  & \multicolumn{3}{c}{5-way 1-shot}              & \multicolumn{3}{c}{5-way 5-shot}              \\
        \cmidrule(lr){1-2}                              \cmidrule(lr){3-5}                              \cmidrule(lr){6-8}
        fw                  & bw                        & Mini   & Tiered & CUB                         & Mini   & Tiered & CUB                         \\
        \midrule
        x                   & x                         & 47.1\% & 52.3\% & 41.8\%                      & 62.9\% & 68.3\% & 59.3\%                      \\
        $\bm{g}(\bm{w}_l)$  & x                         & 48.2\% & 53.6\% & 41.2\%                      & 63.6\% & 69.0\% & 59.0\%                      \\
        $\bm{x}_{l-1}$      & x                         & 37.4\% & 37.9\% & 35.4\%                      & 47.9\% & 49.3\% & 42.5\%                      \\
        x                   & $\bm{g}(\bm{w}_l)$        & 48.0\% & 53.0\% & 42.6\%                      & 64.0\% & 67.8\% & 59.9\%                      \\
        x                   & $\bm{g}(\bm{y}_{l})$      & 22.8\% & 21.1\% & 20.6\%                      & 20.7\% & 21.0\% & 20.4\%                      \\
    \bottomrule
    \end{tabular}
    \begin{tablenotes}
        \footnotesize
        \item
        x: no forward/backward (sparse) meta attention, \ie $\bm{\gamma}_l^{\mathrm{fw}}=1$ or $\bm{\gamma}_l^{\mathrm{bw}}=1$.
    \end{tablenotes}
\end{table}